**Automated Approach for Computer Vision-based Vehicle Movement Classification at Traffic Intersections**

**Udita Jana**
PhD Student, Department of Civil Engineering
Indian Institute of Technology Kanpur, Kanpur, India-208016
Email: uditajana20@iitk.ac.in

**Jyoti Prakash Das Karmakar**
Undergraduate Student, Department of Civil Engineering
Indian Institute of Technology Kanpur, Kanpur, India-208016
Email: jyotipro@iitk.ac.in

**Pranamesh Chakraborty (Corresponding Author)**
Assistant Professor, Department of Civil Engineering
Indian Institute of Technology Kanpur, Kanpur, India-208016
Email: pranames@iitk.ac.in

**Tingting Huang**
Data Scientist
ETALYC Inc., Ames, Iowa, USA-50010
Email: tingting.huang@etalyc.com

**Dave Ness**
Civil Engineer II, City of Dubuque
Dubuque, Iowa, USA-52001
Email: dness@cityofdubuque.org

**Duane Ritcher**
Traffic Engineer, City of Dubuque
Dubuque, Iowa, USA-52001
Email: drichter@cityofdubuque.org

**Anuj Sharma**
Associate Professor, Department of Civil, Construction, and Environmental Engineering
Iowa State University, Ames, Iowa, USA-50011
Email: anujs@iastate.edu

Word Count: 6552 words + 3 table (250 words per table) = 7302 words

*Submitted [August 1, 2021]*

## ABSTRACT

Movement specific vehicle classification and counting at traffic intersections is a crucial component for various traffic management activities. In this context, with recent advancements in computer-vision based techniques, cameras have emerged as a reliable data source for extracting vehicular trajectories from traffic scenes. However, classifying these trajectories by movement type is quite challenging as characteristics of motion trajectories obtained this way vary depending on camera calibrations. Although some existing methods have addressed such classification tasks with decent accuracies, the performance of these methods significantly relied on manual specification of several regions of interest. In this study, we proposed an automated classification method for movement specific classification (such as right-turn, left-turn and through movements) of vision-based vehicle trajectories. Our classification framework identifies different movement patterns observed in a traffic scene using an unsupervised hierarchical clustering technique. Thereafter a similarity-based assignment strategy is adopted to assign incoming vehicle trajectories to identified movement groups. A new similarity measure was designed to overcome the inherent shortcomings of vision-based trajectories. Experimental results demonstrated the effectiveness of the proposed classification approach and its ability to adapt to different traffic scenarios without any manual intervention.

**INTRODUCTION**

Traffic safety is one of the major transportation concerns in the world and in USA. Traffic intersections are one of the critical hotspots for crashes and fatalities. In 2019, out of the 36,096 traffic fatalities, approximately 28% were related to traffic intersections and junctions alone (*1*). This can be attributed due to the complex interactions between vehicles and pedestrians that lead to conflict points and potential accidents hotspots. Moreover, traffic intersections are major bottlenecks of the traffic system and a primary source of traffic delays too. Therefore, traffic intersection monitoring and management are the crucial steps for improving traffic safety and mobility.

One of the first steps of smart traffic intersection monitoring is the efficient collection of traffic data, which includes total vehicle counts in the intersection and classification of turning movements and through movement for the different intersection approaches. Turning and through movement counts are essential for traffic analysis activities like signal-timing optimization and congestion management. These movement specific vehicle counts often guide the decision-making process followed by transportation agencies for tackling complex traffic management problems. As a result, there is a significant demand for such movement count data.

Inductive loop detectors, radar sensors, and bluetooth devices are the common sources that provide traffic count data in general (*2-4*). Inductive loop detectors can provide reliable traffic movement counts. However, they are intrusive, and their installation or repair work is time-consuming and can require stopping traffic movements too. However, the other two alternatives are regarded advantageous as they are non-intrusive, not limited by traffic scenarios, and can provide accurate information about vehicle trajectories. Recently, computer-vision based techniques aided by deep-learning methods have shown promising potential in object detection and classification tasks (*5,6*). Therefore, we can also produce similar vehicle trajectory information from existing traffic surveillance cameras.

Trajectory data extracted from vision-based systems record vehicle's locations with respect to image coordinates. Owing to which the appearance of motion trajectories obtained from vision sensors can change depending on their extrinsic calibrations. Therefore, proposing a particular algorithm for movement classification of vision-based trajectories is a difficult task. Furthermore, the incomplete trajectories, identity switches caused by missed detections and severe occlusions pose a significant challenge at classification. While existing detection-augmentation techniques (*7*) can reduce such problems, they are not totally avoidable.

Several studies on vehicle movement prediction and traffic incident detection also utilized similar vision-based trajectories. In those studies, mainly supervised (*8*) and semi-supervised (*9*) approaches were undertaken for analyzing the vehicle trajectories. Undoubtedly supervised methods can provide better accuracies, but these approaches are not scalable for the movement classification task of our interest. In this regard, the AI City Challenge (*10*) focused on such video-based movement classification. While different strategies were adopted for efficient and effective classification purposes, all of them were guided by relevant features identified manually for individual traffic scenes. The manual effort required here although negligible, is an essential component of these classification schemes, owing to which these approaches cannot readily adapt to different traffic scenes and have limited application in large scale city-wide deployment.

This paper proposes an automated method for computer-vision based movement classification using an unsupervised learning approach. The proposed approach consists of 4 distinct steps, the first being detection of the stopping location for vehicles in intersection approaches and then clustering the vehicle trajectories into movement clusters using unsupervised clustering. Then, the clusters are used to extract modelling trajectories for each movement, which can be compared to the incoming trajectories to classify them accordingly. The major advantage of the proposed approach is that it is an automated method, which can be applied to any traffic intersection approach to perform movement classification, thereby easily scalable to large city-wide applications with minimal manual intervention. The following section gives an overview of previous research on video-based vehicle trajectory classification. The third section specifies the details of our proposed classification algorithm. Section 4 contains description of the dataset used in this study followed by experimental results in Section 5. The final section provides a summary of the paper and briefly outlines the future works.

## RELATED WORK

The first step towards vision-based movement classification is trajectory formation using object detection and tracking algorithms. In the deep learning era, object detection and tracking made tremendous progress both in terms of accuracy and efficiency. CNN algorithms like Faster R-CNN *(11)*, Mask R-CNN *(12)* has achieved state-of-the-art accuracies in object detection tasks. But detection modules like YOLO *(13)*, SSD *(14)*, RetinaNet *(15)* are more efficient and thereby preferred for robust real-time object detection. Various research works are going on to have a better trade-off between accuracy and efficiency. Tracking after detection is a popular strategy adopted for multi object tracking (MOT). Outputs obtained from object detection algorithms across video frames are used in the tracking phase which generates the trajectories of the objects. These tracking algorithms provide spatial as well as temporal information of vehicular movement within the video frames. SORT *(16)*, DeepSORT *(17)* are some of the popular alternatives that provide decent accuracies in MOT problems.

Existing studies that addressed vehicle movement classification tasks from video data can be broadly classified into three categories Line-crossing, Zone-traversal, Trajectory similarity-based classification. The first one follows a line-crossing based approach *(18,19)* where a virtual entry and exit line pair is considered for identifying each movement of Interest (MOI) inside a prespecified region of interest (ROI). This method is not very effective for classifying incomplete trajectories resulting from identity switches or occlusions. Studies belonging to the second category *(20-22)* are similar to the first one but instead of line pairs, several virtual zones (either entrance or both entry-exit zone pairs) are manually selected depending on traffic scenarios. Although these studies produced better accuracies compared to line-crossing based methods, they were not very efficient for unstructured driving environments. The third group of studies *(23,24)* classified vehicle movements following a similarity-based assignment. Representative trajectories for different MOIs were manually chosen for this purpose. These similarity-based methods were found to be much efficient for classifying incomplete vehicle tracks. But they are not scalable when it comes to adapting to various complex traffic scenarios.

Unsupervised methods are generally preferred for increasing scene-adaptability. In this regard, hierarchical clustering-based techniques have been used for classifying vehicular trajectories *(25,26)*. Hausdorff distance between trajectory pairs has been predominantly used as similarity measure for these clustering techniques *(27)*. Traditional Hausdorff distance has certain drawbacks for comparing trajectories of unequal length as it is sensitive to noise and doesn't account for the direction of ordered data pairs. Various improvements *(28,29)* have been proposed to get rid of noise sensitivity and incorporate directional features to the traditional Hausdorff distance. Therefore, unsupervised techniques can be a good alternative towards automated movement classifications.

## METHODOLOGY

The first step towards video-based vehicular movement classification is to extract vehicle trajectories which involves object detection and tracking process. A significant amount of past research has addressed the vision-based detection-tracking problem which is already highlighted in the previous section. Therefore, our study mainly focused on the classification of vehicular trajectories to the corresponding movement types (i.e., through, right-turn, and left-turn). For obtaining the vehicle trajectories, state-of-the-art detection and tracking algorithms have been utilized, details of which are provided in the data description section.

As discussed in the previous section, earlier attempts in this vision-based movement classification task are efficient. But all these methods significantly depend on manual annotations for specifying entry and exit regions of each movement type. Since, this study gears toward obtaining automated movement-specific vehicle count at traffic intersections, we focused on unsupervised methods for addressing such classification tasks.

**Proposed Classification Algorithm**

The proposed movement classification algorithm comprises four stages. At first, 'Stopbar' is located near each intersection approach. In the next stage, vehicle trajectories are clustered according to their movement type which is then followed by modelling trajectory selection. In the final stage, each incoming vehicle's trajectory is compared to modelling trajectories by pre-defined similarity measure and is classified accordingly.

*Stopbar Identification*

This is the first stage of the proposed algorithm. Depending on the coverage of the vision-based detection system, vehicles may be detected and tracked from a much earlier stage even before their arrival near turning locations. Thus, the detection-tracking algorithms generate a significant amount of additional information for each incoming vehicle trajectory. Such additional information is irrelevant for the classification purpose. Due to this, inclusion of the additional information adversely affects the performance of the clustering algorithms and makes the clustering stage more time-consuming and cumbersome. To avoid these complications, the concept of stopbar which was introduced by Santiago-Chaparro et al., *(30)* for radar-based vehicle trajectory analysis has been adapted and applied in this study for vision-based stopbar identification.

For placing the stopbar, first, stopped locations were extracted from tracking outputs obtained for a particular site. In this study, vehicle positions with no displacement in consecutive two video frames were chosen as the stopped location. A horizontal line passing through the 50th percentile value of y-coordinates of these stopped locations are then marked as the stopbar. A visual representation of how stopbars are located from raw trajectory data is shown in **Figure 1**. **Figure 1a** shows all trajectory points obtained from detection and tracking at a sample intersection approach, and **Figure 1b** shows the "stopped locations" for all trajectories as red points and the corresponding "stopbar" line. The vehicle trajectory points shown in figure are with respect to video frame coordinates.

After locating the stopbar (denoted by $Y = Y_{sl}$), only the relevant portions of vehicle trajectories present below the stopbar i.e., in $Y \geq Y_{sl}$ region is extracted and stored as a valid trajectory set $[T_{vs}]$. This valid trajectory set $[T_{vs}]$ is further used in the clustering phase. Since the cameras capturing the traffic movements are facing towards the intersection approach, $Y \geq Y_{sl}$ part or the region below the stopbar in **Figure 1b** ($Y = 0$ marks the top edge of the video frame) is chosen in this case.

Therefore, 'Stopbar' defined in this stage doesn't necessarily align with the lane markings (i.e., stop-lines) or stop signs placed at the intersection legs. Rather, it provides a general understanding of where vehicles are diverging at each incoming approach which is useful for identifying relevant patterns of different movement types. The next stage involves identifying the movement clusters from the trajectories extracted after the stopbar.

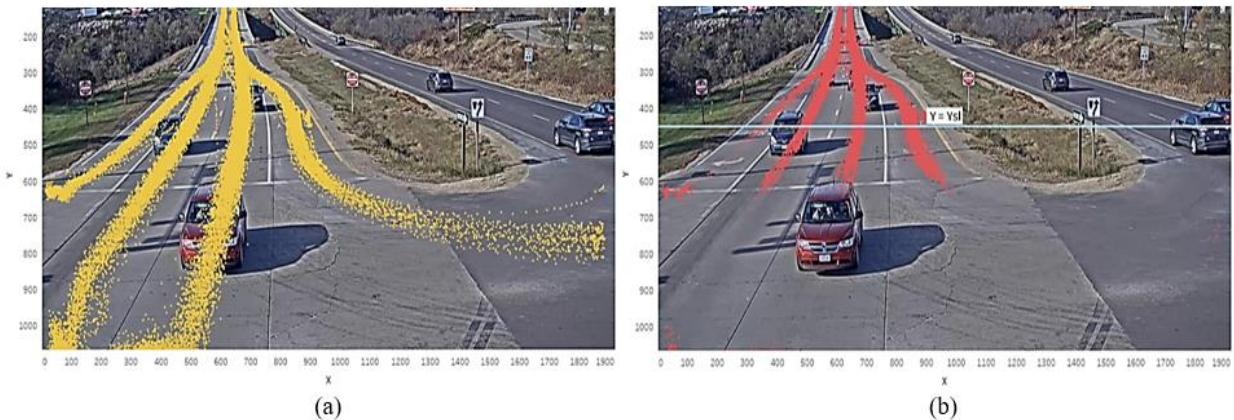

(a)                                                                                      (b)

**Figure 1**. **Stopbar identification process from vehicle trajectories at a sample intersection approach: (a) trajectory points (b) stopped locations (red points) and the stopbar line**

*Identifying movement clusters*

Choice of clustering algorithm and choice of similarity measure are the two major factors that influence the classification process while using unsupervised methods. Clustering algorithms like hierarchical clustering and Density-Based Spatial Clustering of Applications with Noise (DBSCAN) have been mainly used in previous studies for trajectory analysis purpose *(31-33)*. DBSCAN algorithm which has shown promising potential in outlier or anomalous trajectory detection requires the setting of two hyperparameters (epsilon and min points). Selecting appropriate values of these hyperparameters provides site-specific solutions and will result in low-scene adaptability. On the other hand, hierarchical clustering only requires the number of clusters as input, which makes it more suitable for the movement classification task of our interest.

<u>Clustering algorithm</u>- Owing to the reasons stated above, we have used agglomerative clustering algorithm for identifying the movement clusters.

<u>Similarity measure</u>- After the selection of the clustering algorithm, the next critical aspect of trajectory clustering is the choice of similarity measure. In this study, three factors have been considered while designing the similarity measure.

a) Distance similarity- At traffic intersections, vehicles moving in a particular direction generally traverse through a fixed region. Therefore, one of the important aspects while movement classification is to check the spatial proximity between two trajectories. Traditional Hausdorff distance, which has been widely used in the literature for this purpose *(27,28)* returns the maximum of directed Hausdorff distance between a trajectory pair. Since our intended classification task deals with trajectories of dissimilar length and incomplete trajectories, the reverse condition (i.e., minimum value) was adopted for this study. Therefore, the distance similarity between two trajectories '*i*' and '*j*', denoted by $D_S(i,j)$, is defined as,

$$D_S(i,j) = \min{(d_H(i,j), d_H(j,i))} \qquad (1)$$

where, $d_H(P,Q)$ is the directed Hausdorff distance from trajectory '*P*' to '*Q*'. It is the maximum value among minimum distances observed between points of trajectory '*P*' to points of trajectory '*Q*'.

$$d_H(P,Q) = \max_{p \in P}\left\{\min_{q \in Q} d(p,q)\right\} \qquad (2)$$

Here, '*p*', '*q*' stands for data points (vehicle positions in this case) of trajectory '*P*' and '*Q*' respectively. $d(p,q)$ is the Euclidean distance between p and q.

min value used for incomplete trajectory problem.

b) Angular similarity- For traffic intersections, length of obtained vehicle trajectories varies significantly depending on their movement direction (right, left, through) and camera coverage. Along with this, the presence of broken or incomplete trajectories also produces vehicle trajectories of unequal length even within a specific movement group. Therefore, considering the distance similarity alone is not sufficient for such movement classification purposes.

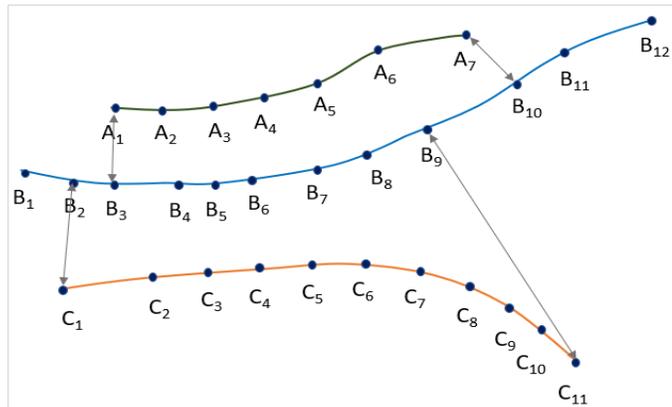

**Figure 2. Illustration for defining angle similarity**

For estimating directional difference between trajectories where incomplete trajectories of varying length might be present across different movement clusters, the following angle similarity measure $T_s$ has been proposed.

$$A_d(i,j) = angle\ between\ \overrightarrow{V_i}\ and\ \overrightarrow{V_{j-i}}\quad where\ A_d(i,j) \in [0,360°] \qquad (3)$$

Where, $\overrightarrow{V_i}$ denotes the vector pointing from starting position to the end position of vehicle trajectory '$i$'.

And $\overrightarrow{V_{j-i}}$ is the vector pointing from closest position of trajectory '$j$' from the starting location of vehicle trajectory '$i$' to the closest position of trajectory '$j$' from the end location of vehicle trajectory '$i$', e.g., for the three trajectories (A, B, C), considering the closest starting and end positions of trajectory B from trajectories A and C as shown in Figure 3.

$$\overrightarrow{V_{B-C}} = \langle B_9[x] - B_2[x], B_9[y] - B_2[y]\rangle$$

here $B_z[*]$ denotes the image coordinates of the $z^{th}$ position in the trajectory B. Similarly, for trajectories A and B,

$$\overrightarrow{V_{B-A}} = \langle B_{10}[x] - B_3[x], B_{10}[y] - B_3[y]\rangle$$

also, for trajectories of different patterns and lengths $\overrightarrow{V_{j-i}} \neq \overrightarrow{V_{i-j}}$

hence, the angle similarity $T_s$ is defined as,

$$T_s(i,j) = \begin{cases} A_d(i,j) & where\ L(i) \leq L(j) \\ A_d(j,i) & where\ L(i) > L(j) \end{cases} \qquad (4)$$

Where, $L(i)$ is the length of trajectory '$i$'. For example, for trajectory 'B' shown in Figure 3,

$$L(B) = \sqrt{(B_{12}[x] - B_1[x])^2 + (B_{12}[y] - B_1[y])^2}$$

c) Proximity of end locations- For vision-based detection systems depending on their extrinsic calibration, two adjacent movement groups might appear as very similar if only the distance and angle similarities are observed as defined above. For example, this can be observed in cases when trajectories right-turning lane and through-movement where they can appear close to each other. However, the right-turn movements can get out of the camera coverage sooner than the through movements, which will lead to different end locations points for right-turn and through movements, as shown in **Figure 1a**. Therefore, when different movement streams are densely located within the video frame, using the previous two similarity aspects might lead to misclassification. The end positions of vehicles, turning (left or right) or proceeding through an intersection are significantly different from each other (unless inflicted with incomplete trajectory problem) and thereby can serve as an additional factor for improving the classification process. Since this study aims to classify trajectories obtained from vision-based sensors, the end locations of vehicular trajectories were also taken into consideration.

In this step, first rear distance $[D_R]$ is computed.

$$D_R(a,b) = \sqrt{(a_{end}[x] - b_{end}[x])^2 + (a_{end}[y] - b_{end}[y])^2} \qquad (5)$$

here, $a_{end}[*]$ denotes the coordinates of the end location of trajectory '$a$'.

The final proximity factor $[P_E]$ is estimated as,

$$P_E(i,j) = \begin{cases} \frac{T_s(i,j)}{3.6} \times \{D_R(i,j) - D_S(i,j)\} & where,\ D_R(i,j) \geq D_S(i,j) \\ \frac{T_s(i,j)}{3.6} \times \{D_R(i,j) - D_S(i,j)\} & where,\ D_R(i,j) < D_S(i,j)\ and\ T_s(i,j) \leq 15 \\ 0 & where,\ D_R(i,j) < D_S(i,j)\ and\ T_s(i,j) > 15 \end{cases} \qquad (6)$$

The final similarity measure $[S(i,j)]$ to be used in the clustering phase is chosen as,

$$S(i,j) = [\{w_1 \times D_S(i,j)\} + \{w_2 \times T_s(i,j)\} + \{w_3 \times P_E(i,j)\}] \qquad (7)$$

A lower value of $S(i,j)$ therefore indicates higher similarities between the two trajectories. As seen from the above formulations, the final proximity factor will increase $S(i,j)$ when the trajectories belong to two different movement groups. Also, this increment is proportional to the angle similarity. As a result, for incomplete trajectories which follow the same movement (although $D_R \geq D_S$ ), this increment will be negligible.

At traffic intersections where more than one lane is dedicated for moving in a certain direction, multiple vehicular streams are observed under a specific movement type. For these cases, rear distance $D_R$ can be smaller than the distance similarity $D_S$ within a movement group (**Figure 1a**). In such scenarios, the proximity factor will decrease the final estimate of $S(i,j)$ thereby reducing dissimilarities between spatially separated vehicular streams within each movement type.

<u>Linkage Rule</u>- To scale dissimilarities between input trajectories, a linkage criterion is to be chosen for hierarchical clustering. Existing linkage rules include minimum distance or single linkage, maximum distance or complete linkage and unweighted average. The following two considerations governed the choice of linkage criteria at this stage.

- o At traffic intersections, multiple vehicular streams can be observed for each movement group.
- o The number of vehicle trajectories observed for different movement groups is not always similar. This also applies to different vehicular streams within each movement group. Owing to which the spread of vehicular streams as obtained from vision-based sensors varies irrespective of their movement direction.

Therefore, the shortest distance or single linkage criterion can be more suitable for this clustering phase. The linkage distance between two clusters $d(g_i, g_j)$ is obtained as

$$d(g_i, g_j) = \min(S(a, b)) \, \forall \, a \in g_i, b \in g_j \qquad (8)$$

Where $g_i, g_j$ are two different movement groups obtained from clustering and a, b are the vehicular trajectories present in those respective movement groups. The movement clusters obtained in this stage are further used to extract modelling trajectories for each movement of interest.

*Modelling trajectory selection*

Although classifying vehicular trajectories by clustering is an automated approach, this isn't preferable for analyzing a large number of vehicular trajectories (as such cases will create a significant calculation overhead). To overcome this problem, the concept of similarity-based assignment has been used in this study. This concept, which was also utilized in several manual based movement classification methods *(23,24)*, demands the specification of modelling trajectories. Such trajectories represent the general trend of how vehicles move while turning or proceeding through the intersection. Hence, in this stage, one or more trajectories are to be chosen from each identified movement cluster as modelling trajectories.

For an automated selection of modelling trajectories, the identified movement clusters are again grouped into one or more sub-clusters depending on the number of lanes dedicated for each movement direction. A similar hierarchical clustering process, as discussed above in the "identifying movement clusters" step, is followed for this purpose. However, unlike the previous clustering stage, the objective here is to find intra-cluster lane-specific movement patterns. Therefore, in this clustering phase, the proximity factor is not considered for obtaining the final similarity measure (as it might separate out incomplete trajectories from full-length ones). Also, vehicle trajectories are not always uniformly spaced within each identified cluster and sometimes even within lane-specific movement groups. As a result, proceeding with single linkage rule might group trajectories passing through different lanes into one cluster. Since our aim here is to obtain lane-specific trajectory groups, single linkage criterion is replaced with average linkage.

<u>Selection criterion</u>- Appropriate selection of modelling trajectories is crucial for the final classification stage. Ideally, the central trajectory should be chosen for this purpose, but since video-based trajectories are curbed with broken and incomplete trajectory problems, the central trajectory found from sub-clustering might include incomplete trajectories, which won't serve the intended purpose of modelling trajectory. Therefore, the longest trajectories of the identified sub-clusters are selected as modelling trajectories for that specific movement of interest.

*Movement Assignment*

In this stage, each incoming vehicle trajectory is compared with the modelling trajectories as identified from the previous stage. The modelling trajectory which is most similar (i.e., resulting in minimum $S(i, j)$ value) to the input vehicle trajectory is identified and the corresponding movement type is assigned to the incoming vehicle id.

Similarities of the input trajectory with the modelling trajectories are computed using predefined similarity measure as shown in **Equation 7**. The proximity factor for end locations is omitted from the final similarity term, for the calculation purpose. This is done for the following reason.

The longest trajectories representing each movement group is used as reference or modelling trajectory. So, the modelling trajectory is not necessarily the central trajectory of each identified movement cluster. Therefore, considering the end-proximity factor might classify the incomplete trajectories of one movement type with an adjacent movement class, the end location of which is more similar to those incomplete trajectories belonging to the first movement type.

Following this similarity-based movement assignment, the movement directions for all vehicle trajectories included in the testing phase is identified, which is further checked with the actual movement directions to evaluate the performance of our proposed algorithm.

**Evaluation Metrics**

For specifying the evaluation metrics, we first define 4 cases – True Positive (TP), True Negative (TN), False Positive (FP) and False Negative (FN). TP is when an algorithm classifies a movement correctly. TN is when an algorithm rejects certain movement classes correctly. FP is when the algorithm wrongly assigns a movement class and FN is when that algorithm wrongly rejects certain movement classes.

| | Classes | Predicted | | |
|---|---|---|---|---|
| | | Right | Left | Through |
| **Actual** | Right | TN | FP | TN |
| | Left | FN | TP | FN |
| | Through | TN | FP | TN |

**Figure 3. Confusion matrix with class Left as reference**

Accuracy is calculated as the ratio of total number of correct classifications to total number of trajectories to be classified. It is obtained as

$$Accuracy = \frac{no.of\ elements\ correctly\ classified\ by\ model\ [i.e. \sum diagonal\ entries\ of\ confusion\ matrix]}{all\ elements\ classified\ by\ the\ model} \quad (9)$$

Balanced accuracy is a performance metric parameter that is used to handle an imbalanced dataset. It is the macro-average of recall scores or true positive rate (TPR) per class
Where,

$$TPR = \frac{TP}{TP+FN} \quad (10)$$

The overall balanced accuracy is calculated as the macro-average of balanced accuracy of all classes.

$$(Balanced\ Accuracy)_{overall} = \frac{\sum(Balanced\ Acuuracy)_i}{Number\ of\ classes} \quad (11)$$

F1 score is another parameter which is considered to get performance for those cases where the number of classes are imbalanced. It is the harmonic mean of precision and recall. These can be expressed as,

$$Precision = \frac{TP}{TP+FP} \quad (12) \qquad Recall = \frac{TP}{TP+FN} \quad (13)$$

For the overall F1 score, we take the macro-average for all classes

$$(F1\ Score)_{overall} = \frac{\sum(F1\ Score)_i}{Number\ of\ classes} \quad (14)$$

**Overview of methods used for comparison**

To understand how our proposed algorithm compares with alternative approaches, three existing methods designed for video-based vehicle trajectory classification have been chosen. The first alternative taken into consideration is line-based classification *(18)*. This is a manual approach that requires specification of virtual entry-exit lines for each movement of interest. The second one proposed by Liu. et. al., *(23)* is a semi-automated approach that looks into shape-similarities of trajectory pairs. This approach follows a rule-based assignment strategy after manual selection of modelling trajectories. Along with these two methods, an automated approach was also included for comparison purposes. In this method proposed by Hao. et. al., *(29)* hierarchical clustering was adopted with length scale directive Hausdorff (LSD Hausdorff) distance as similarity measure. A brief summary of all methods included in the comparison stage is presented in **Table 1**.

**TABLE 1 Methods Used for Comparison**

| Line based | Shape similarity based | LSD Hausdroff | Proposed Method |
|---|---|---|---|
| ▪ Virtual entry-exit line pair manually identified for each MOI.<br>▪ Vehicle crossing these line-pairs are assigned the corresponding MOI. | ▪ Set of modelling trajectories are manually selected for each MOI from line-based method.<br>▪ Distance and directional similarities with modelling trajectories are considered.<br>▪ Assigned to movement where highest similarity is observed given similarity values are within predefined limits. | ▪ Hierarchical clustering with average linkage criteria used for movement clustering.<br>▪ LSD Hausdorff distance used as similarity measure. | ▪ Stop-line identification.<br>▪ Hierarchical clustering with single linkage for movement clustering.<br>▪ Similarity measure defined in section 4 used.<br>▪ Sub-clustering phase for modelling trajectory selection.<br>▪ Similarity based movement assignment. |

**DESCRIPTION OF DATA**

Video data used for the current study were collected from two traffic intersections located in Dubuque, Iowa. The videos were obtained from cameras each facing towards a single inbound intersection approach. In total, this dataset contains 8 videos (each of 4-hours duration), capturing vehicle movements from all four inbound approaches of these two intersections.

For this study, we adopted DeepStream SDK *(34)* to process multiple video streams. We used a ResNet18 model that is pretrained by traffic camera data. It supports the classes of car, bicycle, person, and road sign. We also adopted the tracker based on discriminative correlation filter (DCF). Our detection-tracking module generates a sequence of bounding box coordinates for each unique vehicle observed in the video stream, thereby forming a trajectory.

A visualization of the vehicle trajectory data points obtained from the above-mentioned detection-tracking framework is shown in Figure 3. At each inbound approach, the $1^{st}$ 30 minutes video has been used for training purpose. Vehicle trajectories obtained from the remaining 3.5 hours of video duration has been used for final evaluation. The comparative performance of our proposed algorithm and the three chosen existing methods on this dataset is presented in the next section.

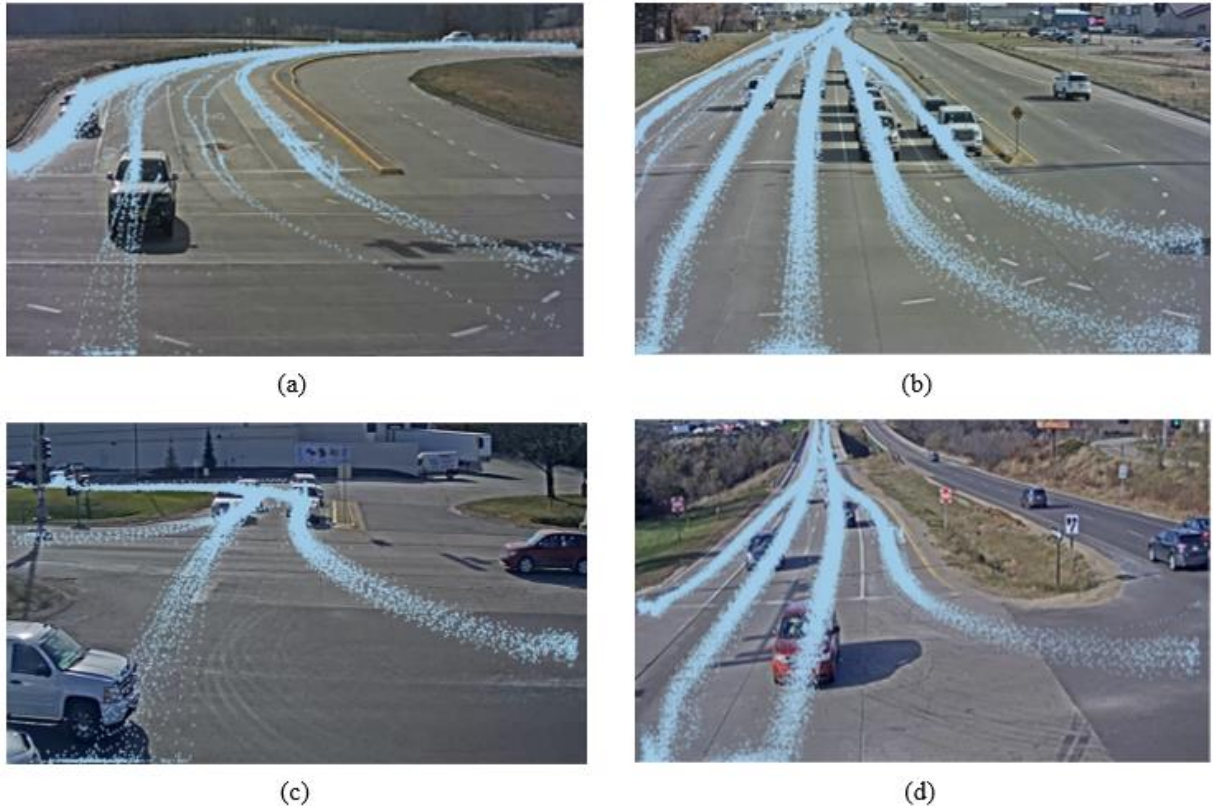

(a)

(b)

(c)

(d)

**Figure 4. Sample Raw Trajectory Data Obtained from Two Intersection Approaches:**
**(a), (b): NB and WB approach of intersection 1; (c), (d) NB and EB approach of intersection 2**

**RESULTS**

As described in the methodology section, our proposed method identifies the modelling trajectories representative of each movement group following a three-step training phase. These trajectories are then used in the final movement assignment stage. A visual representation of the results obtained at these three steps is shown in **Figure 5**. **Figure 5a** shows the stopbar location identified from trajectories of a sample intersection approach, while **Figure 5b** shows the movement cluster identification from trajectories obtained beyond the stop bar. **Figure 5c** shows the third step of the proposed approach, modelling trajectory selection. Note here, the raw trajectory dataset (as seen in **Figure 5a**) used here are extracted from the first 30 min video stream only.

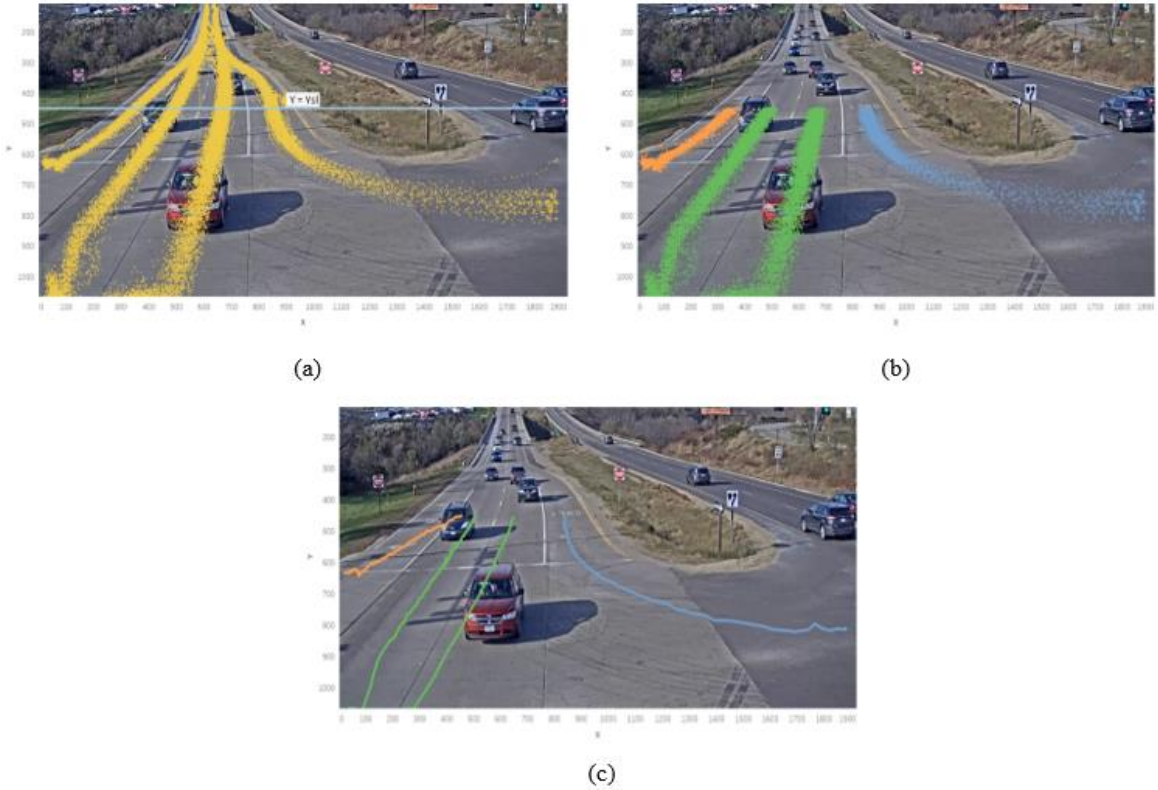

**Figure 5. Sample results from different stages of the proposed approach:
(a) stopbar identification, (b) movement clusters, (c) modelling trajectory**

To understand how the different classification models (listed in Table 1) perform across different traffic scenarios, model performances were grouped into two categories as NB-SB and EB-WB. Average of individual model performances in north-bound and south-bound approaches are listed in the first category, while the second category provides the average of individual model performances observed in east-bound and west-bound approaches.

Such categories were formed as for north-bound and south-bound approaches mostly single vehicular streams were observed for each movement group and these trajectories were sparsely located within the video-frame (see **Figure 4a, 4c**), whereas, in the other category i.e., in the east and west-bound approaches (**see Figure 4b, 4d**) vehicular streams were densely located and multiple lanes were dedicated to one or more movement group.

The performance of different methods across the two specific categories as evaluated on the test dataset is shown in **Table 2**. Accuracy, balanced accuracy and F1 score values listed here are computed using equations 9-14.

**TABLE 2 Comparison of the proposed algorithm with the baseline methods**

| Method | Accuracy (%) | | balanced accuracy (%) | | F1 score | |
|---|---|---|---|---|---|---|
| | NB-SB | EB-WB | NB-SB | EB-WB | NB-SB | EB-WB |
| Line based | 95.76 | 96.10 | 96.86 | 97.61 | 0.98 | 0.98 |
| Shape similarity based | 89.88 | 99.01 | 85.48 | 98.34 | 0.94 | 0.99 |
| LSD Hausdorff | 89.36 | 56.87 | 89.21 | 69.71 | 0.90 | 0.64 |
| **Proposed approach** | **99.64** | **99.80** | **99.42** | **99.90** | **1.00** | **1.00** |

The comparative performance of different approaches considered are highlighted below-

a. Performance of line-crossing based method didn't vary across different sites since for this method virtual entry-exit line pairs for each MOI are carefully placed observing movement patterns at each location. The only source of error was incomplete trajectories that didn't cross any of those pre-specified line pairs and hence were not classified.

b. For the shape-similarity based approach, achieved accuracies were comparatively lower in NB-SB locations. This is due to the fixed distance and angle thresholds considered in this method. For NB-SB locations in the dataset, mainly single vehicular streams are observed for each MOI, and one or more full-length central trajectories are chosen as modelling trajectories. Therefore, vehicle trajectories that follow a different path while moving in a certain direction remain unclassified if they exceed either of the two thresholds although it has the lowest similarity with the modelling trajectories of its movement direction.

   For EB-WB locations closely placed movement groups are observed and mostly multiple vehicular streams are present within each movement group. Therefore, the percentages of unclassified trajectories arising from fixed threshold considerations are significantly less in these locations.

c. The third method, LSD Hausdorff, although automated was only able to classify movement patterns with large spatial separation. As a result, it achieved accuracies around 90% for NB-SB locations. Whereas for EB-WB locations where vehicle trajectories were densely located within the video frame this method was prone to misclassification and the classification accuracies reduced significantly.

d. As seen from Table 3, the proposed algorithm in this study performed well although different intersection locations of the used dataset and achieved comparable or higher accuracies for both NB-SB and EB-WB approaches. Among the four different methods used for comparison, this is the only automated approach that isn't limited by traffic scenarios. The classification process was least influenced by the presence of broken or incomplete trajectories which is an unavoidable aspect of vision-based trajectories.

In addition to the vehicle trajectory data, the proposed algorithm utilizes movement-specific lane-level information for automated selection of modelling trajectories. To understand how our classification method will perform in absence of such movement-specific lane level information, the proposed classification algorithm has been also assessed with single modelling trajectory selection criteria (i.e., the longest trajectory from each movement group is chosen in this case without any sub-clustering. A comparison of the algorithm performance under these two different modelling trajectory selection criteria is shown in **Table 3**.

**TABLE 3 Effect of Modelling Trajectory selection on Proposed Algorithm Performance**

| Location | Modelling Trajectory Choice | Accuracy (%) | balanced accuracy (%) | F1 score |
|---|---|---|---|---|
| North bound South bound | Lane-specific | 99.64 | 99.42 | 1.00 |
| | Single | 99.71 | 99.54 | 1.00 |
| East bound West bound | Lane-specific | 99.80 | 99.90 | 1.00 |
| | Single | 88.09 | 95.47 | 0.90 |

It can be observed that for EB-WB locations, where densely located vehicle trajectories have to be classified, selecting multiple modelling trajectories according to dedicated number of lanes significantly improves the algorithm performance. On the other hand, in NB-SB locations where movement groups are separated by a large spatial margin, the choice of modelling trajectories had negligible influence over the algorithm performance. Nonetheless, the proposed modelling trajectory selection criteria aided our classification approach to achieve high accuracy percentages without any scene-specific distance and angle threshold consideration.

## CONCLUSIONS

In this study, we tried to address the movement classification task for vehicle trajectories at traffic intersections obtained from vision-based sensors. A fully automated method has been proposed for this purpose considering the inherent shortcomings of video-based detection systems. Further, the choice of similarity measure carefully designed for non-uniform motion trajectories and the use of lane-specific modelling trajectories helped our classification approach adapt to different traffic scenarios. It has been observed that the appearance of movement patterns within the video frame has negligible influence on algorithm performance. As a result, the proposed method achieved high accuracies and F1-score close to 1 for all the eight intersection approaches used for evaluation. Experimental results on this dataset show that in comparison to existing alternatives, our method has similar or better performance for such movement classification tasks, without requiring any manual intervention and thereby can be scalable for implementation at large city-wide or district-wide levels.

The current stopbar identification process, (which is a crucial part of our proposed classification framework) is designed for trajectories obtained from a single intersection approach. In future, we'll extend this stopbar identification part for applying in traffic scenes where multiple inbound approaches are visible within a single video frame. We also intend to explore deep-learning based object detection models for automatic extraction of lane-level or stop-line information relevant for movement classification purpose.

## AUTHOR CONTRIBUTIONS

The authors confirm contribution to the paper as follows: study conception and design: U. Jana, P. Chakraborty, D. Ness, D. Ritcher, A. Sharma; data collection: D. Ness, D. Ritcher, A. Sharma, T. Huang; analysis and interpretation of results: U. Jana, J. P. D. Karmakar, P. Chakraborty, T. Huang; draft manuscript preparation: U. Jana, J. P. D. Karmakar, P. Chakraborty, T. Huang. All authors reviewed the results and approved the final version of the manuscript.

## ACKNOWLEDGEMENTS

Our research results are based upon work supported by the Science and Engineering Research Board (SERB) under Sanction Order No. SRB/2020/001708. Any opinions, findings, and conclusions or recommendations expressed in this material are those of the author(s) and do not necessarily reflect the views of the SERB.